\documentclass[letterpaper, 10 pt, conference]{ieeeconf}  

\IEEEoverridecommandlockouts                              

\usepackage{graphicx}
\usepackage{subcaption} 
\usepackage{mathptmx}
\usepackage{amsmath}
\usepackage{amssymb}
\usepackage{hyperref}
\usepackage{bm}

\title{\LARGE \bf
See, Learn, Assist: Safe and Self-Paced Robotic Rehabilitation via
Video-Based Learning from Demonstration
}

\author{Ali Alabbas$^{1}$, Camillo Murgia$^{1}$, Joanne Regan$^{2}$, and Philip Long$^{1}$%
\thanks{$^{1}$Authors are with the Faculty of Mechanical Engineering, Atlantic Technological University, Galway, Ireland. Email: {\tt\small ali.alabbas@research.atu.ie}}%
\thanks{$^{2}$Author is with the Department of Health and Nutritional Science, Atlantic Technological University, Sligo, Ireland.}%
\thanks{This work is supported by the HEA's TU RISE programme, co-financed by the Government of Ireland and the European Union via the European Regional Development Fund (ERDF) under the Southern, Eastern \& Midland, and Northern \& Western Regional Programmes 2021–27.}%
}

\begin{document}

\maketitle
\thispagestyle{empty}
\pagestyle{empty}

\begin{abstract}

In this paper, we propose a novel framework that allows therapists to teach robot-assisted rehabilitation exercises remotely via RGB-D video. Our system encodes demonstrations as 6-DoF body-centric trajectories using Cartesian Dynamic Movement Primitives (DMPs), ensuring accurate posture-independent spatial generalisation across diverse patient anatomies. Crucially, we execute these trajectories through a decoupled hybrid control architecture that constructs a spatially compliant virtual tunnel, paired with an effort-based temporal dilation mechanism. This architecture is applied to three distinct rehabilitation modalities: Passive, Active-Assisted, and Active-Resistive, by dynamically linking the exercise's execution phase to the patient's tangential force contribution. To guarantee safety, a Gaussian Mixture Regression (GMR) model is learned on-the-fly from the patient’s own limb. This allows the detection of abnormal interaction forces and, if necessary, reverses the trajectory to prevent injury. Experimental validation demonstrates the system's precision, achieving an average trajectory reproduction error of $3.7$~cm and a range of motion (ROM) error of $5.5^\circ$. Furthermore, dynamic interaction trials confirm that the controller successfully enforces effort-based progression while maintaining strict spatial path adherence against human disturbances.
\end{abstract}

\section{INTRODUCTION}

The demand for rehabilitation is steadily increasing due to an aging and more active population. By 2050, the number of people aged over 65 is projected to reach 129.8 million in Europe \cite{EUstats}.
Additionally, the rising prevalence of degenerative diseases is placing significant strain on existing rehabilitation resources~\cite{Ju2023}.

Robotic-assisted rehabilitation offers a promising solution to deliver repeatable, high-quality therapy while reducing clinical workload. However, traditional robotic solutions face significant trade-offs: exoskeletons offer precise joint control but are often bulky and expensive \cite{banyai2024robotics, lee2020comparisons}, while soft robots provide compliance but lack the force capacity required for full limb rehabilitation \cite{tanczak2025soft, morris2023state}. Furthermore, end-effector systems, though robust, operate primarily in task space, making it difficult to enforce anatomically correct joint movements \cite{paolucci2021robotic, hogan1992manus, wu2024compact}. Even isokinetic dynamometers like the Biodex \cite{tankevicius2013test}, a standard system for clinical joint-level assessment, require tedious physical reconfiguration and are typically restricted to a single degree of freedom.

\begin{figure}[!tbp]
  \centering
  \includegraphics[width=\linewidth]{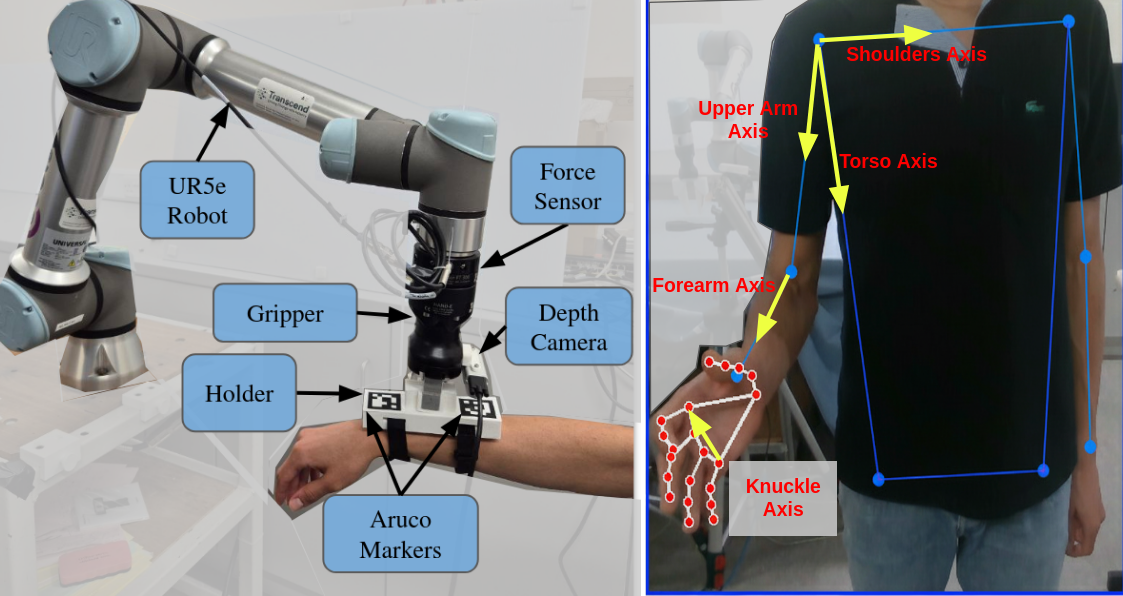}
  \caption{Hardware architecture of the system (left) and body-centric frames constructed from skeleton keypoints (right).}
  \label{fig:setup}
\end{figure}
We aim to overcome this complexity by leveraging collaborative robots (cobots). Cobots provide a safe, versatile, and compliant alternative for close-proximity rehabilitation. Typically, cobot programming for rehabilitation relies on kinesthetic Learning from Demonstration (LfD) \cite{auta2025robot}, which requires the physical co-location of the therapist and the robot. While recent advances in video-based imitation learning offer a pathway to reduce this burden via remote RGB-D recordings \cite{wang2023mimicplay, li2024okami, qin2022dexmv}, translating these purely spatial, kinematic trajectories into safe, force-interactive physical rehabilitation remains an open challenge.

To address this gap, this paper introduces a novel video-based LfD framework capable of teaching a cobot physical, multi-modal upper-limb rehabilitation exercises directly from a therapist's remote video demonstration. The primary contributions of this work are as follows:
\begin{enumerate}
    \item Body-Centric Video LfD: A markerless perception pipeline that encodes therapist demonstrations as 6-DoF Cartesian DMPs within a body-centric reference frame. This ensures accurate, anatomically posture-independent scaled spatial generalisation across diverse patient geometries.
    \item Multi-Modal Hybrid Control: A controller, coupled with an effort-based temporal dilation law, that constructs a virtual tunnel around the exercise trajectory to prevent unsafe deviations. Moreover, by linking the execution speed to the patient's tangential effort, three execution modalities are enabled: Passive, Active-Assisted, and Active-Resistive.
    \item Force-Adaptive Reversibility: A personalized safety mechanism utilizing a GMR model calibrated on-the-fly to the patient's limb, monitoring interaction forces and triggering a smooth trajectory reversal if abnormal forces are detected.
\end{enumerate}
The paper is organized as follows. Section ~\ref{Sec:related-work} reviews related literature. Section~\ref{Sec:sys-arch} presents the system architecture. Section~\ref{Sec:exp-val} reports the experimental validation and pilot studies. Finally, Section~\ref{Sec:Dis-work} discusses the results and highlights directions for future work.

\section{Related Work} \label{Sec:related-work}
\subsection{Robot-Assisted Rehabilitation and Interaction Control}
Optimizing physical human-robot interaction (pHRI) is critical for promoting neuroplasticity in robotic rehabilitation. To encourage active participation, recent Assist-as-Needed (AAN) paradigms rely heavily on variable impedance and admittance control. For instance, \cite{zhang2024research} and \cite{manoharan2025user} use real-time parameter estimation and Bayesian optimization to dynamically adjust robot stiffness based on user intent, while~\cite{chen2024adaptive} integrates adaptive admittance with backstepping control for compliant, accurate trajectory tracking.

Beyond baseline assistance, varying stages of recovery demand multi-modal operation and spatial adaptation. Jutinico et al. \cite{jutinico2017impedance} employ robust Markovian impedance control to safely transition between passive, resistive, and fixed modes. Similarly, \cite{xu2023rehabilitation} simultaneously adapts spatial trajectories and assistance levels based on patient performance. While these strategies excel at force-level adaptation, their reliance on rigidly pre-programmed paths or wearable sensors motivates the need for intuitive, vision-based trajectory generation integrated with force-adaptive execution.

\subsection{Vision-Based Task Transfer in pHRI}
Teaching robots via video demonstration is rapidly emerging as a scalable alternative to kinesthetic teaching for pHRI. For instance, combining visual tracking with sEMG and force data allows DMPs to model human-like motion and variable impedance \cite{zhang2025human}, while one-shot imitation learning frameworks seamlessly map human hand trajectories from a single video to robot end-effectors using DMPs for skill adaptation \cite{lu2025visual}. Furthermore, Vision-Language-Action policies optimize dynamic collaborations by mapping video demonstrations to generalised end-effector trajectories \cite{kareer2025emergence}, and frameworks like OKAMI achieve single-video skill transfer via object-aware motion retargeting \cite{li2024okami}. The precision of this spatial transfer is clinically vital; recent studies highlight that pHRI directly mediates postural errors during stroke recovery \cite{chen2025physical}. Building on these spatial transfer principles, our framework utilizes YOLO-based skeletal tracking and Cartesian DMPs to directly extract, scale, and reproduce full-limb therapeutic motions.

\subsection{Safety Mechanisms and Anomaly Handling in pHRI}

Dynamic human-robot interaction requires robust safety mechanisms to manage spatial deviations and involuntary anomalies like muscle spasms. To enforce safe workspaces, ~\cite{miao2024integrating} bounds user motion using learned repulsive force fields. To detect these sudden anomalies, \cite{luo2024research} demonstrates that multi-sensor fusion significantly improves intent detection accuracy. Consequently, recent frameworks utilize such multi-sensor data \cite{sun2022intelligent} to safely switch between passive and active training modes, or fuzzy impedance control \cite{pan2013safety} to safely halt the robot during emergencies. However, abrupt halting disrupts therapeutic flow. To avoid this, Escarabajal et al. \cite{escarabajal2023imitation} trigger smooth trajectory reversals when interaction forces violate a baseline safety corridor learned from the patient’s healthy limb. Our framework guides motion along a virtual tunnel with low-gain orthogonal compliance,  while a personalized GMR model monitors interaction forces to enable smooth reversal in case of anomaly detection.

\section{Methodology} \label{Sec:sys-arch}
Our framework consists of three core components: a perception module for markerless motion capture, a learning module for body-centric trajectory encoding, and a force-adaptive execution module for safe exercise reproduction.

\subsection{Perception Module}

The Perception Module extracts the therapist’s demonstration from the raw RGB-D stream. We employ YOLOv8-pose \cite{yolov8_ultralytics} for robust body skeleton detection under challenging conditions (e.g., occlusion, varying lighting) \cite{maji2022yolo} and complement it with Mediapipe Hand \cite{lugaresi2019mediapipe} for fine-grained hand landmarks. Depth data is used to project these 2D keypoints into 3D space. While this yields the 3D position vectors for each keypoint, the orientation information required to fully capture the therapeutic motion is absent. Consequently, we construct local 6-DoF joint frames based on the geometric relationships between the acquired keypoints.

To ensure the recorded motion is encoded relative to the user, we construct a body-centric reference frame. The first frame's base orientation is derived using three stable landmarks: the two shoulders and the right hip. This orientation is then anchored at the shoulder joint to create the local coordinate system. By defining the motion within this frame, the system decouples the arm's movement from the patient's trunk posture, ensuring the exercise remains biomechanically accurate even if the patient leans or shifts position during therapy, as shown in Fig.~\ref{fig:setup}.

However, detecting forearm axial rotation (pronation and supination) is challenging, as simple wrist and elbow keypoints often fail to capture the twisting motion. We resolve this by tracking a secondary orientation vector across the hand's knuckles, which is then cross-referenced with the primary wrist–elbow alignment axis. This allows the reliable differentiation between a supinated and pronated grasp. Finally, to mitigate the noise associated with keypoint estimation and depth camera tracking, a power moving-average filter is applied to all trajectories. This smooths the input data, yielding a stable sequence of 6-DoF poses for the targeted limb, which are then passed to the learning module.

\subsection{Learning Module}
Post-demonstration, the learning module encodes the 6-DoF motion using Cartesian DMPs \cite{ude2014orientation}. DMPs approximate the trajectories with weighted Gaussian basis functions, offering low computational complexity suitable for real-time applications. We have selected Cartesian DMPs for their ability to encode full 6-DoF trajectories, ensuring the accurate reproduction of both position and orientation essential for rehabilitation exercises. Moreover, since DMPs can generalise spatially, the encoded movement can be dynamically stretched or compressed to fit a specific patient's limb length, which serves to generalise across different body sizes. Lastly, DMPs can compensate for the patient's range of motion as start/goal positions can be updated without distorting the fundamental shape of the exercise.

Crucially, the system learns this trajectory within the local body-centric frame (relative to the shoulder anchor) rather than the global camera frame. By decoupling the motion from the environment, the learned skill becomes invariant to the patient's position. This ensures that the robot executes the movement relative to the patient's anatomy, preserving the biomechanical intent of the therapist, even if the patient is seated differently or shifts posture between sessions.

\subsection{Multi-Modal Execution and Safety Module}

This module translates the spatially generalised DMP trajectories into safe pHRI. To accommodate varying stages of motor recovery, we implement a decoupled control architecture that separates spatial path enforcement from temporal execution progress.

\subsubsection{Effort-Modulated Virtual Tunnel Control Architecture} \label{Subsection: virtual_tunnel}
To ensure anatomically correct motion while enabling dynamic interaction, the patient's limb motion is guided along a virtual tunnel defined around the 6-DoF learned trajectory, using a hybrid control architecture that modulates the DMP's progression speed based on the patient's own effort. At each control step, the unit tangent direction of the trajectory is computed as:
\begin{equation}
\bm{u}_{\mathrm{t}} = \bm{v}_{\mathrm{ref}}/||\bm{v}_{\mathrm{ref}}||,    
\end{equation}
where $\bm{v}_{\mathrm{ref}}$ is the 3D linear reference velocity generated by the learned Cartesian DMPs. Since the canonical phase $s$ is monotonically non-increasing under the modulation law defined in Equation~\eqref{eq:phase_dynamics}, the direction of $\bm{v}_{\mathrm{ref}}$ coincides with the geometric tangent of the learned path at the current phase, independent of the instantaneous progression speed $\dot{s}$. The force measured by the robot's force/torque sensor, $\bm{f}_{\mathrm{ex}}$, is projected onto this tangent to extract the tangential effort:
\begin{equation}
 f_t = \bm{f}_{\mathrm{ex}} \cdot \bm{u}_{\mathrm{t}}.   
\end{equation}
The remaining orthogonal component is isolated as:
\begin{equation}
 \bm{f}_{\mathrm{o}} = \bm{f}_{\mathrm{ex}} - f_t \bm{u}_{\mathrm{t}}.   
\end{equation}
Along the tangential direction, the objective is to couple patient effort directly to exercise progression, letting the patient set the pace of the movement. This is achieved by modulating the phase of the DMP canonical system with the measured tangential force:
\begin{equation}\label{eq:phase_dynamics}
 \tau \dot{s} = - \max(\epsilon_{\mathrm{min}}, \epsilon + \gamma f_t) s,    
\end{equation}

where $\tau$ is the temporal scaling, $\gamma$ is the sensitivity gain, $\epsilon$ is the baseline progression rate, and $\epsilon_{min}$ is a strictly positive constant preventing complete trajectory stalling should the patient pause. By tuning $\gamma$ and $\epsilon$ relative to the expected range of $f_t$, different execution modalities can be obtained, ranging from near-passive guidance to fully patient-paced progression. 

For the orthogonal axes, to emulate the guidance of a therapist without rigidly constraining the patient to the exercise path, a low-gain admittance law governs motion along the orthogonal axes:
\begin{equation}
\bm{v}_{\mathrm{wall}} = \mathbf{A}_{\mathrm{stiff}} \bm{f}_{\mathrm{o}}    
\end{equation}
where $\mathbf{A}_{\mathrm{stiff}} \in \mathbb{R}^{3 \times 3}$ is a diagonal admittance gain matrix. The gain $\mathbf{A}_{\mathrm{stiff}}$ is chosen to be small, so that a given patient effort produces substantially less orthogonal motion than tangential progression. This creates a directional bias favoring tangential progression over orthogonal deviation, without imposing a hard constraint. While this zeroth-order admittance bounds the rate of orthogonal deviation, it does not cap accumulated displacement under sustained force. Enforcing a strict anatomical corridor via an additional restoring stiffness is left as an extension for future work.

\subsubsection{Multi-Modal Operation}
To adapt patient engagement to different recovery stages, we use the force-coupled temporal modulation defined in Equation~\eqref{eq:phase_dynamics}. By adjusting $\gamma$ and $\epsilon$, the system scales across three rehabilitation modalities:\\
\textbf{Passive Mode}: Designed for the acute recovery phase when the patient cannot initiate movement. In this mode, we set $\gamma=0$ and $\epsilon=\alpha$, where $\alpha \in (0, 1]$ is the temporal scaling factor. Progression is entirely robot-driven and independent of interaction force, so the canonical system evolves purely as a function of time, gently driving the passive limb along the spatial tunnel at a therapist-defined speed.\\
\textbf{Active-Assisted Mode} (AAN): Designed for patients capable of actively engaging, but too weak to complete the movement independently. The sensitivity gain $\gamma$ is activated and tuned to achieve the required patient engagement (lower values need more effort to progress). $\epsilon$ value is kept low for the robot to provide a slow, continuous baseline movement. The trajectory accelerates when $f_t> 0$, and drops to a minimal speed if the patient stops or resists ($f_t\leq 0$). This ensures self-paced, active motor engagement while providing necessary robotic support.\\
\textbf{Active-Resistive Mode}: For late-stage therapy targeting strength and endurance, $\epsilon$ is reduced toward $\epsilon_{\mathrm{min}}$, while $\gamma$ is significantly reduced to force the patient to exert sustained, significant effort to advance the DMP phase and complete the exercise.

\begin{figure}[!tbp]
    \centering
    
    \begin{subfigure}[t]{0.5\linewidth}
        \centering
        \includegraphics[width=\linewidth]{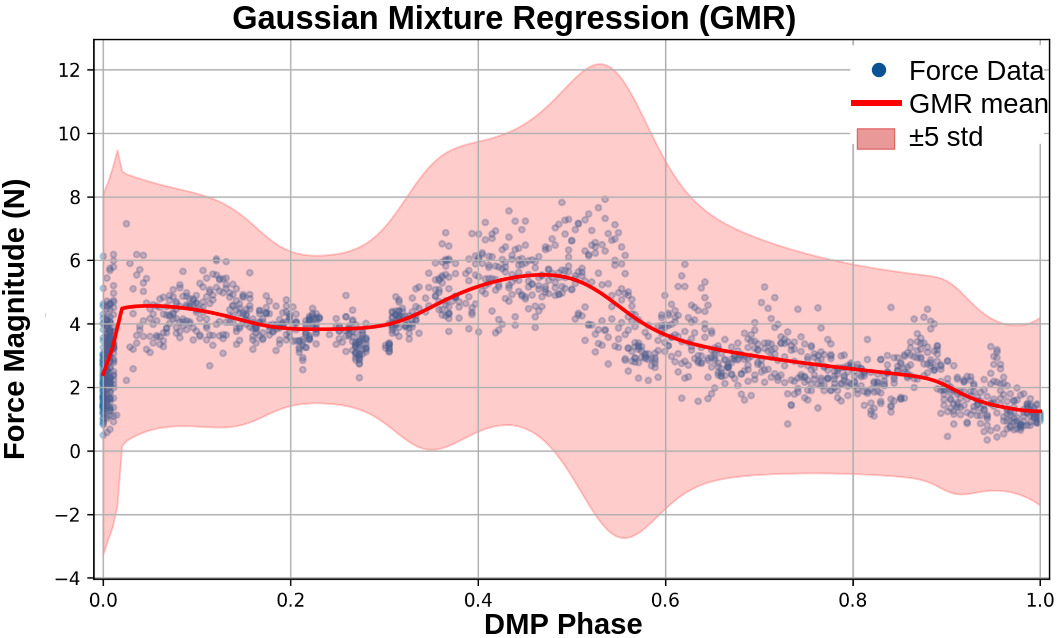}
        \label{fig:GMRleft}
    \end{subfigure}
    \hfill
    \begin{subfigure}[t]{0.48\linewidth}
        \centering
        \includegraphics[width=\linewidth]{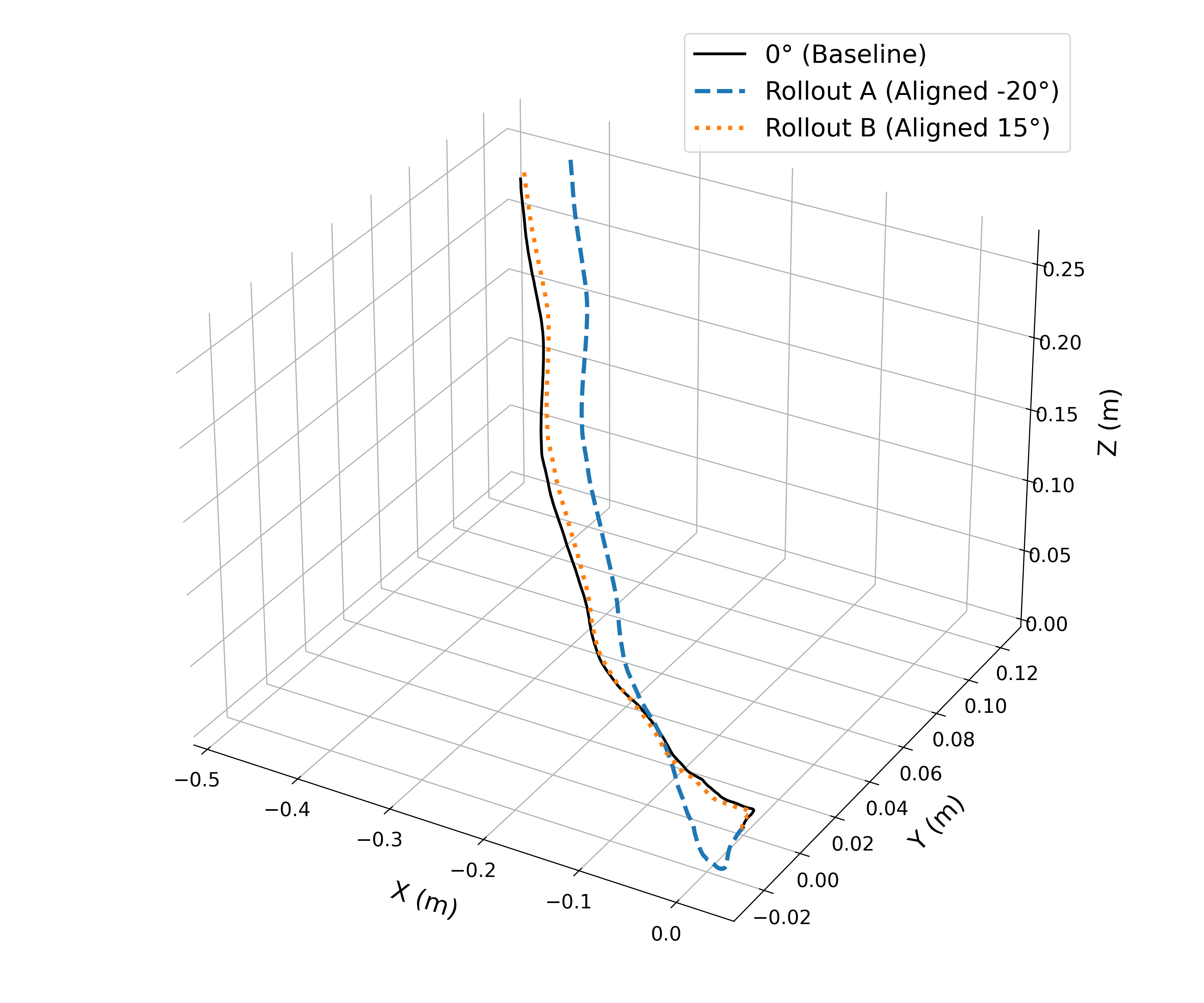}
        \label{fig:right}
    \end{subfigure}
    \caption{A GMR model learned from a subject's limb during elbow flexion-extension exercise (left). The safety corridor is defined as a $\pm5$ standard deviation around the mean expected force (in red). (Right) Exercise trajectories during shoulder abduction-adduction executed at varying trunk rotation angles. The trajectories are aligned at their starting points to demonstrate the high fidelity of the body-centric encoding.}
    \label{fig:two_eps}
\end{figure}

\subsubsection{Force-Adaptive Safety and Reversibility} \label{Sub_sec:GMR_safety}
To ensure patient safety and comfort, a personalized baseline of normal interaction forces is established prior to therapy. Determining this baseline depends on the selected rehabilitation modality. For the passive mode, this baseline is inferred from the patient's healthy limb, similar to~\cite{escarabajal2023imitation}. The robot guides the healthy limb through the trajectory, recording forces against the DMP phase variable $s$ to train a Gaussian Mixture Model (GMM). Conversely, for the Active-Assisted and Active-Resistive modes, the baseline is learned directly on the targeted, impaired limb. The robot is set to a gravity-compensated state along the tangential axis, passively collecting the patient's exerted forces and the DMP phase. The GMM links the forces to the phase variable of the DMP, rather than pure time,  accommodating the variable execution speeds inherent to patient-driven movement.

During rehabilitation, GMR is used to predict the expected mean force and variance at each phase as shown in Fig.~\ref{fig:two_eps} (left). This probabilistic estimate is used to construct a dynamic safety corridor, defined as $\pm5$ standard deviations around the predicted mean, inspired by the work in \cite{escarabajal2023imitation}. If the measured interaction force violates the safety corridor, indicating spasticity, fatigue, or the limits of the patient's range of motion, the system triggers a smooth reversal along the previously traversed trajectory. This ensures that motion returns through regions already verified to be safe. Once the force returns to normal bounds, the robot automatically resumes forward motion. This reversible behavior enables self-paced rehabilitation while preventing pain or excessive loading.

\section{Experimental Validation} \label{Sec:exp-val} 
To evaluate the efficacy and safety, we conduct a set of experiments using the hardware setup shown in Fig.~\ref{fig:setup}.  The system comprises a UR5e cobot, a Robotiq Hand-E gripper, and two D435 RGB-D cameras: a tripod-mounted camera to record the expert demonstration, and a wrist-mounted camera (on the UR5e) to observe the patient during execution. Additionally, the patient wears a custom 3D-printed wrist handle to provide a grasping interface for the robot with two ArUco markers for fine position estimation of the grasping point. We assume that the maximum applied interaction force does not exceed $40$~N.

First, we validate the spatial accuracy and anatomical scaling across different subjects of the vision-based LfD pipeline. Secondly, we evaluate the system's dynamic control capabilities, specifically the enforcement of the virtual tunnel and the effort-based temporal dilation. Finally, we test the GMR-based safety reversibility mechanism. All experiments received institutional ethics approval, and trials were supervised by a researcher. Both the participant and the researcher had access to an emergency stop button to halt the robot's operation at any time.

\subsection{Spatial Accuracy and Anatomical Generalisation}
The objective of this experiment is to verify the end-to-end spatial fidelity, ensuring the robot reproduces the therapeutic intent and generalises across varying patient anthropometries. Under the guidance of an exercise specialist, we evaluated three fundamental upper-limb exercises spanning the sagittal, transverse, and frontal planes of motion, respectively: elbow flexion-extension, shoulder internal-external rotation, and shoulder abduction-adduction. Ground truth motion data was captured using an OptiTrack motion capture system (mocap), which tracked the subjects' skeletal kinematics and recorded the motion of the wrist handle worn by the participants.

\subsubsection{Baseline Reproduction Accuracy}
To validate the accuracy of our system reproduction, we asked a healthy subject to demonstrate the three exercises wearing the Optitrack suit in front of a tripod-mounted camera. The robot then reproduced the exercises on the same subject. Table \ref{tab:reproduction_results} shows the reproduction error between the exercise trajectories and their robot-reproduced ones. The Root Mean Square Error (RMSE) of the reproduction for the three exercises was less than $5$~cm, which proves the system's capability of translating the motion learned from the video demonstration. Moreover, to assess the baseline fidelity of the body-centric encoding, the same subject positioned themselves with a varying angle in front of the robot: facing the robot directly, then with the trunk rotated $-20^\circ$ (clockwise), and with the trunk rotated $+15^\circ$ (counter-clockwise). We used for validation the learned DMPs for shoulder abduction-adduction exercise. Fig. \ref{fig:two_eps} (right) shows the exercise trajectory in the three cases. For better visualization, we aligned the starting point of the three trajectories.  The RMSE between each pair of the three trajectories was less than $1$~cm,  validating that the body-centric motion encoding successfully preserves the biomechanical integrity, independent of the subject's posture.

\begin{table}[!bh]
\centering
\caption{Reproduction RMSE for the three exercises.}
\begin{tabular}{l c}
\hline
\textbf{Exercise} & \textbf{Reproduction Error (cm)} \\
\hline
Elbow flexion–extension & 1.8 \\
Shoulder internal–external rotation & 4.5 \\
Shoulder abduction–adduction & 4.9 \\
\hline
\end{tabular}
\label{tab:reproduction_results}
\end{table}

\subsubsection{Anatomical Scaling}
In physical rehabilitation, the therapeutic intent of an exercise is defined by the joint angular displacement, or ROM. For single-joint exercises, this angular displacement $\theta$ can be approximated by the ratio of the end-effector path distance $D$ to the limb length $L$, such that $ \theta \approx D/L $. An unscaled trajectory would force patients with shorter limbs into over-extension, while leaving patients with longer limbs under-actuated. To mitigate this, our system learns normalized DMPs and then scales them based on the patient's limb length.

\begin{table}[!tbp]
\centering
\caption{Normalized reach ratio ($R$) across different limb lengths: scaled vs. unscaled baseline.}
\resizebox{\columnwidth}{!}{%
    \begin{tabular}{l c c}
    \hline
    \textbf{Limb Length (m)} & \textbf{Path Distance (m)} & \textbf{Normalized Ratio (R)} \\
    \hline
    0.61 & 0.47 & 0.77 \\
    0.55 & 0.43 & 0.78 \\
    0.53 & 0.41 & 0.77 \\
    0.50 & 0.38 & 0.76 \\
    \hline
    0.50 (Unscaled) & 0.47 & 0.94 \\
    \hline
    \end{tabular}%
}
\label{tab:generalization}
\end{table}

We conducted a pilot study with four healthy subjects with upper limb lengths: $0.50$~m, $0.53$~m, $0.55$~m, and $0.61$~m. The subject with the longer limb demonstrated the exercise, and the robot replicated the normalized exercise on the four subjects. We define the Normalized Reach Ratio as $R=D/L$, where a successful DMP scaling must maintain a consistent $R$ across subjects to preserve the target joint angle. During execution, the perception module estimates $L$ from skeletal keypoints and scales the Cartesian DMP output in real-time. As shown in Table \ref{tab:generalization}, the system maintained a highly consistent reach ratio across all participants, exhibiting a deviation of less than $4\%$, which shows the method's capacity to generalise a single reference motion across diverse body types. In contrast, an unscaled baseline test on the participant with the shortest limb resulted in a target overshoot ($R=0.94$ with a deviation of $\approx 22\%$). 

\begin{figure}[!tbp]
  \centering
  \includegraphics[width=\linewidth]{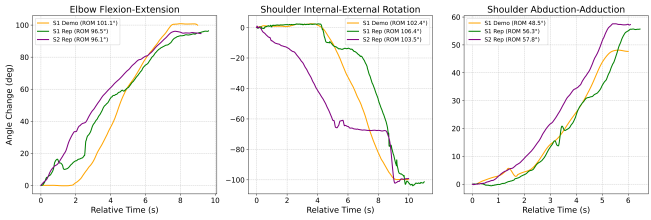}
  \caption{Comparison of time-normalized joint angle trajectories for the three upper-limb exercises. Orange lines show the demonstration (subject 1), while green and purple lines show the robot reproductions for subjects 1 and 2.}
  \label{fig:optitrack_results}
\end{figure}

\subsubsection{Kinematic Fidelity and Range of Motion}
Reference exercises were demonstrated by a subject with a limb length of $0.61$~m and recorded by the mocap system. The robot then reproduced this motion for both the original subject and a new subject with a $0.55$~m arm length to evaluate the scaling fidelity. Fig.~\ref{fig:optitrack_results} illustrates the joint angle trajectories (Flexion/Extension and Abduction/Adduction) recorded during these trials. For visualization, all trajectories are time-normalized to align their execution durations. The average ROM deviation from the original demonstration was approximately $5.5^\circ$, with Shoulder Abduction–Adduction exhibiting the largest deviation of roughly $9^\circ$.

We attribute this error primarily to reduced accuracy in skeletal keypoint detection due to the low reflectivity of the OptiTrack suit, which degraded visual tracking performance, particularly when the arm moved close to the body. Overall, the system maintained good kinematic fidelity and successfully reproduced upper-limb exercises across users with different limb lengths, as the $9^\circ$ worst-case deviation is comparable to the $\approx 8^\circ$ manual measurement error typically observed between therapists \cite{akizuki2016effect}.

\begin{figure}[!tbp]
  \centering
  \includegraphics[width=\linewidth]{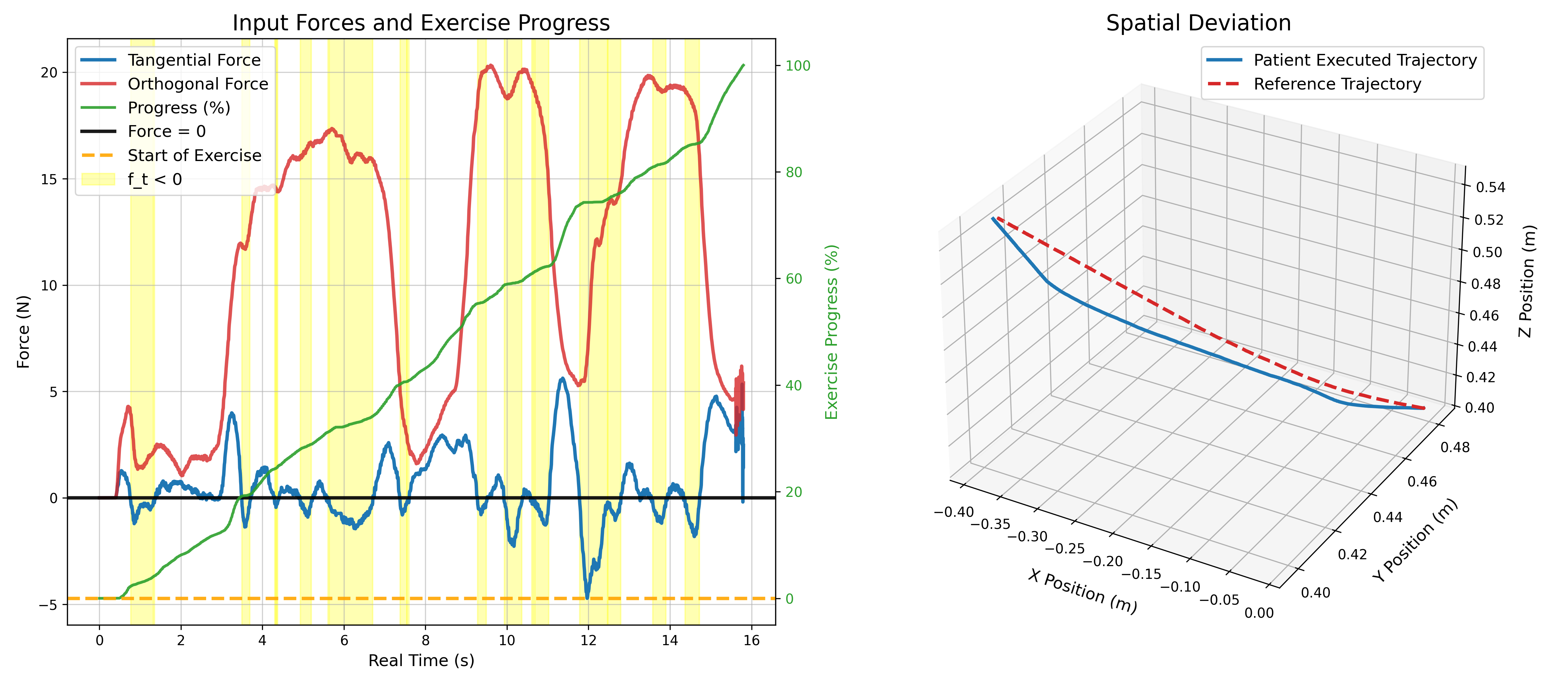}
  \caption{Validation of the virtual tunnel and effort-based temporal dilation. (Left) The subject intentionally applies significant orthogonal forces (in red) while varying tangential effort (in blue). Exercise progress (green) is driven only by $f_t$, slowing when $f_t < 0$. (Right) 3D Cartesian paths confirm that the stiff orthogonal admittance maintains spatial adherence to the reference trajectory despite the significant force applied.}
  \label{fig:virtual_tunnel}
\end{figure}

\subsection{Virtual Tunnel Validation}
Our system guides the limb's motion inside a virtual tunnel learned from the expert's demonstration, as discussed in Section~\ref{Subsection: virtual_tunnel}. To validate the spatial robustness, a healthy participant performed the shoulder rotation exercise in active-assisted mode. To simulate a lack of motor coordination, the participant was asked to push in the orthogonal direction against the tunnel walls and at specific intervals during execution. The stiff orthogonal admittance gain was set to $\mathbf{A}_{\mathrm{stiff}} = 0.001 \mathbf{I}_{3\times3}$, and the temporal dilation parameters were tuned to $\gamma = 0.04$, and $\epsilon = 0.001$. The controller operated at a frequency of $500$~Hz, and the robot's Tool Center Point (TCP) path was recorded relative to the robot's base coordinate frame.

As depicted in Fig.~\ref{fig:virtual_tunnel} (left), despite the subject's orthogonal forces ($||\bm{f}_{\mathrm{o}}||$ in red) being significant, the progress of the exercise remained governed only by the tangential force along the tangential axis. The low-gain orthogonal admittance kept the resulting trajectory deviation small relative to the applied disturbance. To establish a baseline for comparison, we first recorded a reference trajectory by setting the temporal parameters to $\gamma = 0.0$ and $\epsilon = 0.5$, allowing the system to progress autonomously at a constant rate. The maximum recorded spatial deviation from the reference trajectory was $3.7$~cm (Fig.~\ref{fig:virtual_tunnel} (right)), confirming that the decoupled controller effectively biased the motion toward the learned trajectory, limiting the influence of the patient's orthogonal effort while ensuring the patient retains an intuitive feeling of control over the execution speed.

\begin{figure}[!tbp]
  \centering
  \includegraphics[width=\linewidth]{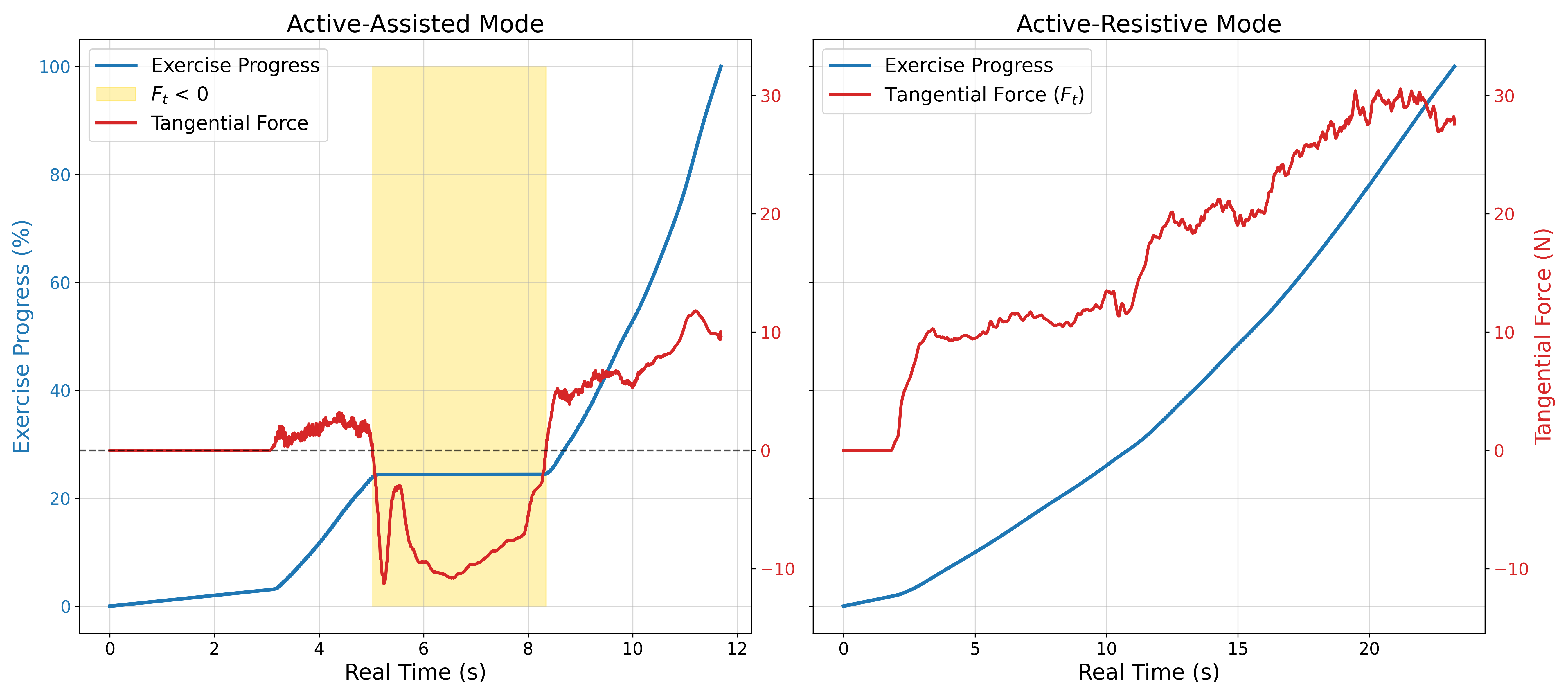}
  \caption{Exercise progress vs patient's delivered force in the active-assisted and active-resistive modes.}
  \label{fig:temporal_dilation}
\end{figure}

\subsection{Effort-Based Dilation and Multi-Modal Execution}
To evaluate the temporal dynamics of the patient-driven execution, we analyze the system's response across the active-assisted and active-resistive modalities. The objective is to validate the force-driven exercise progression governed by the effort-based dilation law in~(\ref{eq:phase_dynamics}), and to demonstrate how the patient intuitively controls the exercise flow through active physical engagement.

\begin{figure*}[!t]
  \centering
  \includegraphics[width=\linewidth, height=0.3\linewidth]{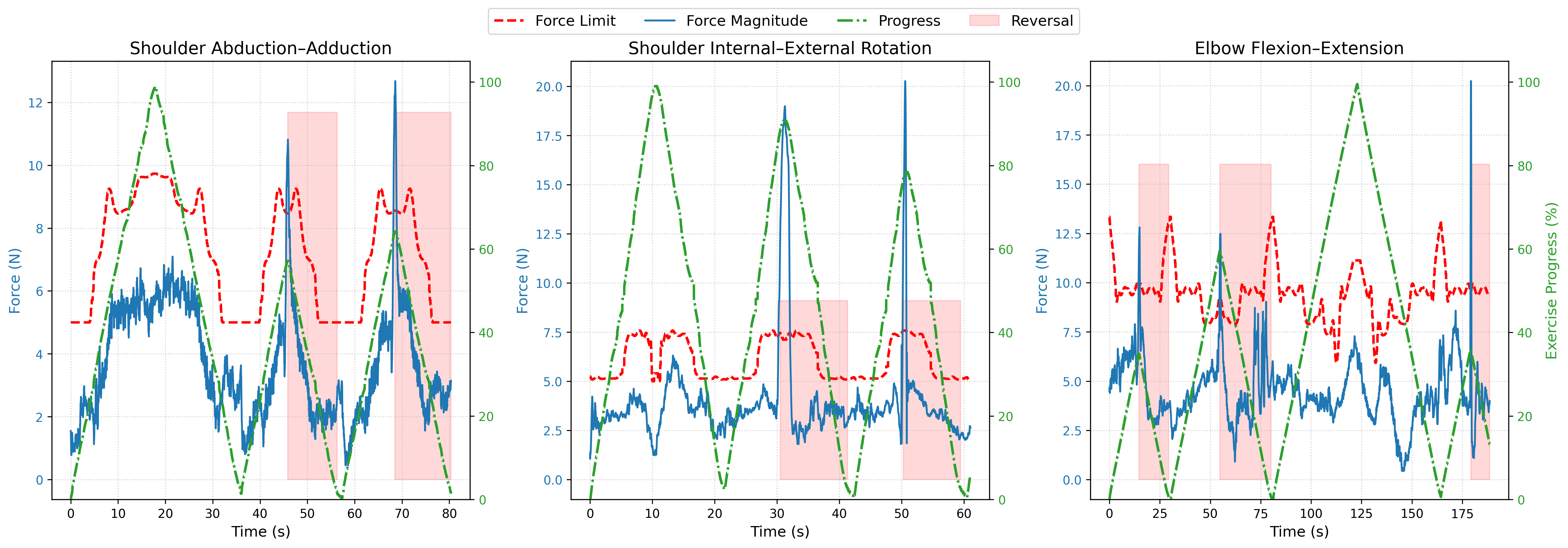}
  \caption{Force monitoring and adaptive reversibility during the three rehabilitation exercises. The green line indicates exercise progress as a percentage of the demonstration. When the measured interaction force (blue line) exceeds the learned safety limit (dashed red) predicted by the GMR model, the system triggers smooth trajectory reversal to maintain patient comfort.}
  \label{fig:trajectory}
\end{figure*}

\subsubsection{Active-Assisted Mode}
To configure the system for a responsive assist state, we set $\gamma = 0.04$ and $\epsilon=0.001$, while $\epsilon_{min} = 0.0001$. The value of $\gamma$ was chosen relative to an estimated maximum tangential effort of a recovering patient in this mode, $f_{t,\mathrm{max}} \approx 25$~N, so that $\gamma \leq \frac{1- \epsilon}{f_{t,\mathrm{max}}}$. This ensures the DMP never runs faster than the demonstrated pace. A subject was asked to execute the exercise with variable levels of engagement to test the controller's responsiveness. As illustrated in Fig.~\ref{fig:temporal_dilation} (left), the exercise progress increases with the applied tangential force. When the participant intentionally stops pushing or applies a resistive force (highlighted zone), the canonical system of the DMP only progresses by the minimal baseline speed, allowing the robot to wait for the patient's renewed effort while eliminating the risk of the subject slacking or being overpowered. Once $f_t>0$ again, the trajectory accelerates toward completion. To filter sensor noise and involuntary disturbances along the tangential axis, we employed an empirically tuned $2$~N deadband. However, adaptive noise rejection and intent recognition will be explored in future work.

\subsubsection{Active-Resistive Mode}
In this mode, we reduced $\gamma = 0.005$ and set $\epsilon = \epsilon_{min} =0.0001$. The participant was asked to push as hard as possible when completing the exercise. As shown in Fig.~\ref{fig:temporal_dilation} (right), the effort required to advance the phase is larger. To ensure a steady progression, the subject must exert and sustain a significantly higher tangential force, gradually climbing to a peak of approximately $30$~N. Consequently, the exercise requires nearly twice as long to complete ($23$~s). This demonstrates the controller's ability to safely provide the resistance needed to build muscular strength and endurance, all while guiding the patient effort within the virtual tunnel.

\subsection{Passivity analysis}

We model the dynamic relationship between the orthogonal interaction force $F_o(s)$ and the realized end-effector velocity $v_o(s)$ as a time-delayed first-order system (where $s$ denotes the Laplace variable here):
\begin{equation}
    Y(s) = \frac{v_o(s)}{F_o(s)} = A_{\mathrm{stiff}} \cdot e^{-s \tau_d} \cdot \frac{1}{T_c s + 1},
\end{equation}
where $\tau_d = 0.002$~s is the communication latency and $T_c \approx 0.01$~s is the UR5e's inner velocity-tracking time constant. Strict passivity requires $\mathrm{Re}\{Y(j\omega)\} \ge 0$, which bounds the total phase lag introduced by the delay and the inner loop to $90^\circ$:
\begin{equation}
    \angle Y(j\omega) = -\omega \tau_d - \arctan(\omega T_c) \ge -90^\circ.
\end{equation}
Under these parameters, the linear model remains passive up to approximately $f_{\text{lin}} \approx 34.4$~Hz.

Beyond this linear analysis, the UR5e's $2\ \mathrm{m/s^2}$ acceleration limit introduces a second, amplitude-dependent constraint. For a sinusoidal input $F(t) = F_{\max}\sin(2\pi f t)$, the commanded velocity requires a peak acceleration of $ a_{peak} = A_{\mathrm{stiff}} \cdot F_{\max} \cdot 2\pi f$. Requiring this to remain below the $2\ \mathrm{m/s^2}$ limit (with $A_{\mathrm{stiff}} = 0.001$) yields a joint stability envelope:
\begin{equation}
    F_{\max} \cdot f \le 318\ \mathrm{N \cdot Hz}.
\end{equation}
Our application involves interaction forces up to $40$~N during intentional motion and frequencies up to $18$~Hz for pathological tremor \cite{bhatia2018consensus}. At $40$~N, the boundary permits frequencies up to $\sim\!8$~Hz, well above the $\sim\!4$~Hz frequency of voluntary human motion. For a maximum tremor frequency of $18$~Hz, the boundary permits force amplitudes of $\sim\!17$~N, comfortably above the low-amplitude nature of neurological tremor. The system therefore avoids acceleration saturation and remains strictly passive across all targeted biomechanical conditions.

\subsection{Force-Adaptive Safety and Reversibility}
To evaluate the GMR-based safety controller, we tested the system's ability to monitor interaction forces dynamically and execute safe trajectory reversals. We established a personalized GMM/GMR model as described in Section ~\ref{Sub_sec:GMR_safety} and shown in Fig. ~\ref{fig:two_eps}.

During the reproduction phase, the participant was instructed to intentionally exert sudden resistive forces to simulate spasticity, pain, or an unexpected restriction in their range of motion. The system was operating in passive mode. As depicted in Fig.~\ref{fig:trajectory}, the system continuously evaluates the real-time interaction force magnitude (blue line) against the dynamic safety limit predicted by the GMR model (dashed red line).

While the interaction forces remain within the expected bounds, the exercise progresses smoothly (indicated by the green progress line). However, the moment the measured force violates the safety corridor, the controller triggers a smooth trajectory reversal (highlighted by the red shaded regions). The robot reverses along the already-traversed, verified-safe path until the abnormal forces return to nominal levels or until reaching the starting point. It then resumes the forward motion. This enables self-paced rehabilitation while ensuring exercise repetition without causing patient discomfort. The reaction time of the system was $\approx 0.3$~s measured from the moment the force violation occurs to the start of the reversal motion. In Active-Assisted and Active-Resistive modes, this latency poses no risk because the patient's tangential effort drives the progression. In the Passive mode, the baseline speed is set to $0.1$~m/s to minimize the displacement during this brief latency. Independent of the GMR-based reversal, a hard safety limit of $50$~N is enforced to immediately halt the robot. Future clinical studies will investigate the effect of this reaction time on patient comfort across different scenarios.

\section{CONCLUSIONS AND FUTURE WORK} \label{Sec:Dis-work}

We presented a novel video-based LfD framework that encodes therapist demonstrations into body-centric 6-DoF trajectories for rehabilitation exercises, ensuring safe execution through GMR-based force modeling and trajectory reversibility. The system delivers therapy across passive, active-assisted, and active-resistive modalities. During active modes, an effort-based temporal dilation mechanism empowers the patient to intuitively control the exercise progression through their applied tangential effort, while passive mode safely guides the passive limb along the reference path. The framework was validated across three exercises encompassing the three anatomical planes of motion. For subjects with varying limb lengths, the vision-based anatomical scaling successfully reproduced the expert’s intended motion with an average positional error of $3.7$~cm. Furthermore, the system maintained high kinematic fidelity, preserving the target ROM with a minimal average deviation of $5.5^\circ$.

Despite these results, several limitations remain. First, while the system successfully identifies discrete grasp orientations, the vision pipeline struggles to capture continuous axial joint rotations, which are inherently difficult for markerless pose estimators and limit the fidelity of rotational exercises. Second, performance degrades under occlusions or suboptimal camera placement, as body-centric frame construction requires continuous visibility of key anatomical landmarks. Third, the system currently relies on a single wrist-mounted holder as the grasping interface, requiring all trajectories to be mapped to the wrist. Finally, the experimental validation is currently limited to a pilot study of four healthy subjects.

While this study demonstrates the technical feasibility and safety of the proposed framework in experiments with healthy participants, clinical trials involving patients with motor impairments are necessary to assess its therapeutic efficacy. Future work will therefore focus on conducting clinical user studies with both physiotherapists and patients. Specifically, we aim to evaluate the physical impact of the system's transient reversal latency on patient comfort. Furthermore, we will explore adaptive noise rejection and intent recognition to better distinguish voluntary effort from pathological tremors. Additionally, we aim to eliminate the manual tuning required for the hybrid controller by integrating Vision-Language Models (VLMs) to interpret high-level therapist instructions. Furthermore, current rehabilitation robotic solutions ensure fixation of the patient’s body segments not involved in the motion, an assumption we also adopted in this study, but we plan to implement continuous real-time patient pose monitoring to adjust the exercise trajectories. Ultimately, we aim to leverage the learned trajectories and force-adaptive interaction data as training inputs for generative models that adapt dynamically to patient ability and recovery progress.

\addtolength{\textheight}{-3cm}   



\begin{thebibliography}{10}
\providecommand{\url}[1]{#1}
\csname url@samestyle\endcsname
\providecommand{\newblock}{\relax}
\providecommand{\bibinfo}[2]{#2}
\providecommand{\BIBentrySTDinterwordspacing}{\spaceskip=0pt\relax}
\providecommand{\BIBentryALTinterwordstretchfactor}{4}
\providecommand{\BIBentryALTinterwordspacing}{\spaceskip=\fontdimen2\font plus
\BIBentryALTinterwordstretchfactor\fontdimen3\font minus \fontdimen4\font\relax}
\providecommand{\BIBforeignlanguage}[2]{{%
\expandafter\ifx\csname l@#1\endcsname\relax
\typeout{** WARNING: IEEEtran.bst: No hyphenation pattern has been}%
\typeout{** loaded for the language `#1'. Using the pattern for}%
\typeout{** the default language instead.}%
\else
\language=\csname l@#1\endcsname
\fi
#2}}
\providecommand{\BIBdecl}{\relax}
\BIBdecl

\bibitem{EUstats}
\BIBentryALTinterwordspacing
``Ageing europe - statictics on population developments,'' 2024. [Online]. Available: \url{https://ec.europa.eu/eurostat/statistics-explained/index.php?title=Ageing_Europe_-_statistics_on_population_developments}
\BIBentrySTDinterwordspacing

\bibitem{Ju2023}
F.~Ju, Y.~Wang, B.~Xie, Y.~Mi, M.~Zhao, and J.~Cao, ``The use of sports rehabilitation robotics to assist in the recovery of physical abilities in elderly patients with degenerative diseases: A literature review,'' \emph{Healthcare (Basel)}, vol.~11, no.~3, p. 326, 2023.

\bibitem{banyai2024robotics}
A.~D. Banyai and C.~Brișan, ``Robotics in physical rehabilitation: Systematic review,'' in \emph{Healthcare}, vol.~12, no.~17.\hskip 1em plus 0.5em minus 0.4em\relax MDPI, 2024, p. 1720.

\bibitem{lee2020comparisons}
S.~H. Lee, G.~Park, D.~Y. Cho, H.~Y. Kim, J.-Y. Lee, S.~Kim, S.-B. Park, and J.-H. Shin, ``Comparisons between end-effector and exoskeleton rehabilitation robots regarding upper extremity function among chronic stroke patients with moderate-to-severe upper limb impairment,'' \emph{Scientific reports}, vol.~10, no.~1, p. 1806, 2020.

\bibitem{tanczak2025soft}
N.~Tanczak, A.~Yurkewich, F.~Missiroli, S.~K. Wee, S.~Kager, H.~Choi, K.-J. Cho, H.~K. Yap, C.~Piazza, L.~Masia \emph{et~al.}, ``Soft robotics in upper limb neurorehabilitation and assistance: current clinical evidence and recommendations,'' \emph{Soft robotics}, vol.~12, no.~3, pp. 303--314, 2025.

\bibitem{morris2023state}
L.~Morris, R.~S. Diteesawat, N.~Rahman, A.~Turton, M.~Cramp, and J.~Rossiter, ``The-state-of-the-art of soft robotics to assist mobility: a review of physiotherapist and patient identified limitations of current lower-limb exoskeletons and the potential soft-robotic solutions,'' \emph{Journal of neuroengineering and rehabilitation}, vol.~20, no.~1, p.~18, 2023.

\bibitem{paolucci2021robotic}
T.~Paolucci, F.~Agostini, M.~Mangone, A.~Bernetti, L.~Pezzi, V.~Liotti, E.~Recubini, C.~Cantarella, R.~G. Bellomo, C.~D’Aurizio \emph{et~al.}, ``Robotic rehabilitation for end-effector device and botulinum toxin in upper limb rehabilitation in chronic post-stroke patients: an integrated rehabilitative approach,'' \emph{Neurological Sciences}, vol.~42, no.~12, pp. 5219--5229, 2021.

\bibitem{hogan1992manus}
N.~Hogan, H.~I. Krebs, J.~Charnnarong, P.~Srikrishna, and A.~Sharon, ``Mit-manus: a workstation for manual therapy and training. i,'' in \emph{[1992] Proceedings IEEE International Workshop on Robot and Human Communication}.\hskip 1em plus 0.5em minus 0.4em\relax IEEE, 1992, pp. 161--165.

\bibitem{wu2024compact}
R.~Wu, M.~Luo, J.~Fan, J.~Ma, N.~Zhang, J.~Li, Q.~Li, F.~Gao, and G.~Dan, ``A compact motorized end-effector for ankle rehabilitation training,'' \emph{Frontiers in Robotics and AI}, vol.~11, p. 1453097, 2024.

\bibitem{tankevicius2013test}
G.~Tankevicius, D.~Lankaite, and A.~Krisciunas, ``Test--retest reliability of biodex system 4 pro for isometric ankle-eversion and-inversion measurement,'' \emph{Journal of sport rehabilitation}, vol.~22, no.~3, pp. 212--215, 2013.

\bibitem{auta2025robot}
I.~A. Auta, A.~Fares, H.~Iwata, and H.~El-Hussieny, ``Robot-assisted upper limb rehabilitation using imitation learning,'' \emph{Journal of Robotics and Control (JRC)}, vol.~6, no.~1, pp. 89--100, 2025.

\bibitem{wang2023mimicplay}
C.~Wang, L.~Fan, J.~Sun, R.~Zhang, L.~Fei-Fei, D.~Xu, Y.~Zhu, and A.~Anandkumar, ``Mimicplay: Long-horizon imitation learning by watching human play,'' \emph{arXiv preprint arXiv:2302.12422}, 2023.

\bibitem{li2024okami}
J.~Li, Y.~Zhu, Y.~Xie, Z.~Jiang, M.~Seo, G.~Pavlakos, and Y.~Zhu, ``Okami: Teaching humanoid robots manipulation skills through single video imitation,'' \emph{arXiv preprint arXiv:2410.11792}, 2024.

\bibitem{qin2022dexmv}
Y.~Qin, Y.-H. Wu, S.~Liu, H.~Jiang, R.~Yang, Y.~Fu, and X.~Wang, ``Dexmv: Imitation learning for dexterous manipulation from human videos,'' in \emph{European Conference on Computer Vision}.\hskip 1em plus 0.5em minus 0.4em\relax Springer, 2022, pp. 570--587.

\bibitem{zhang2024research}
Y.~Zhang, T.~Li, H.~Tao, F.~Liu, B.~Hu, M.~Wu, and H.~Yu, ``Research on adaptive impedance control technology of upper limb rehabilitation robot based on impedance parameter prediction,'' \emph{Frontiers in Bioengineering and Biotechnology}, vol.~11, p. 1332689, 2024.

\bibitem{manoharan2025user}
G.~Manoharan and H.~Lee, ``User-adaptive variable impedance control using bayesian optimization for robot-aided ankle rehabilitation,'' \emph{IEEE Transactions on Neural Systems and Rehabilitation Engineering}, 2025.

\bibitem{chen2024adaptive}
L.~Chen, J.~Huang, Y.~Wang, S.~Guo, M.~Wang, and X.~Guo, ``Adaptive patient-cooperative compliant control of lower limb rehabilitation robot,'' \emph{Biomimetic Intelligence and Robotics}, vol.~4, no.~2, p. 100155, 2024.

\bibitem{jutinico2017impedance}
A.~L. Jutinico, J.~C. Jaimes, F.~M. Escalante, J.~C. Perez-Ibarra, M.~H. Terra, and A.~A. Siqueira, ``Impedance control for robotic rehabilitation: a robust markovian approach,'' \emph{Frontiers in neurorobotics}, vol.~11, p.~43, 2017.

\bibitem{xu2023rehabilitation}
J.~Xu, K.~Huang, T.~Zhang, K.~Cao, A.~Ji, L.~Xu, and Y.~Li, ``A rehabilitation robot control framework with adaptation of training tasks and robotic assistance,'' \emph{Frontiers in bioengineering and biotechnology}, vol.~11, p. 1244550, 2023.

\bibitem{zhang2025human}
H.~Zhang, F.~Peng, and M.~Cai, ``Human--robot variable-impedance skill transfer learning based on dynamic movement primitives and a vision system,'' \emph{Sensors}, vol.~25, no.~18, p. 5630, 2025.

\bibitem{lu2025visual}
S.~Lu, C.~H{\"a}rdtlein, and J.~Schilp, ``Visual imitation learning from one-shot demonstration for multi-step robot pick and place tasks,'' \emph{Scientific Reports}, 2025.

\bibitem{kareer2025emergence}
S.~Kareer, K.~Pertsch, J.~Darpinian, J.~Hoffman, D.~Xu, S.~Levine, C.~Finn, and S.~Nair, ``Emergence of human to robot transfer in vision-language-action models,'' \emph{arXiv preprint arXiv:2512.22414}, 2025.

\bibitem{chen2025physical}
Z.-J. Chen, Y.~Chen, J.~Xu, X.-L. Huang, and C.~He, ``Physical human-robot interaction mediates the association of motor impairment and kinematic performance for poststroke arm rehabilitation,'' \emph{BMC Sports Science, Medicine and Rehabilitation}, vol.~17, no.~1, p. 310, 2025.

\bibitem{miao2024integrating}
Q.~Miao, S.~Min, C.~Wang, and Y.-F. Chen, ``Integrating subject-specific workspace constraint and performance-based control strategy in robot-assisted rehabilitation,'' \emph{Frontiers in neuroscience}, vol.~18, p. 1473755, 2024.

\bibitem{luo2024research}
S.~Luo, Q.~Meng, S.~Li, and H.~Yu, ``Research of intent recognition in rehabilitation robots: a systematic review,'' \emph{Disability and Rehabilitation: Assistive Technology}, vol.~19, no.~4, pp. 1307--1318, 2024.

\bibitem{sun2022intelligent}
P.~Sun, R.~Shan, and S.~Wang, ``An intelligent rehabilitation robot with passive and active direct switching training: improving intelligence and security of human--robot interaction systems,'' \emph{IEEE Robotics \& Automation Magazine}, vol.~30, no.~1, pp. 72--83, 2022.

\bibitem{pan2013safety}
L.~Pan, A.~Song, G.~Xu, H.~Li, H.~Zeng, and B.~Xu, ``Safety supervisory strategy for an upper-limb rehabilitation robot based on impedance control,'' \emph{International Journal of Advanced Robotic Systems}, vol.~10, no.~2, p. 127, 2013.

\bibitem{escarabajal2023imitation}
R.~J. Escarabajal, J.~L. Pulloquinga, P.~Zamora-Ortiz, {\'A}.~Valera, V.~Mata, and M.~Vall{\'e}s, ``Imitation learning-based system for the execution of self-paced robotic-assisted passive rehabilitation exercises,'' \emph{IEEE Robotics and Automation Letters}, vol.~8, no.~7, pp. 4283--4290, 2023.

\bibitem{yolov8_ultralytics}
\BIBentryALTinterwordspacing
G.~Jocher, A.~Chaurasia, and J.~Qiu, ``Ultralytics yolov8,'' 2023. [Online]. Available: \url{https://github.com/ultralytics/ultralytics}
\BIBentrySTDinterwordspacing

\bibitem{maji2022yolo}
D.~Maji, S.~Nagori, M.~Mathew, and D.~Poddar, ``Yolo-pose: Enhancing yolo for multi person pose estimation using object keypoint similarity loss,'' in \emph{Proceedings of the IEEE/CVF conference on computer vision and pattern recognition}, 2022, pp. 2637--2646.

\bibitem{lugaresi2019mediapipe}
C.~Lugaresi, J.~Tang, H.~Nash, C.~McClanahan, E.~Uboweja, M.~Hays, F.~Zhang, C.-L. Chang, M.~G. Yong, J.~Lee \emph{et~al.}, ``Mediapipe: A framework for building perception pipelines,'' \emph{arXiv preprint arXiv:1906.08172}, 2019.

\bibitem{ude2014orientation}
A.~Ude, B.~Nemec, T.~Petri{\'c}, and J.~Morimoto, ``Orientation in cartesian space dynamic movement primitives,'' in \emph{2014 IEEE International Conference on Robotics and Automation (ICRA)}.\hskip 1em plus 0.5em minus 0.4em\relax IEEE, 2014, pp. 2997--3004.

\bibitem{akizuki2016effect}
K.~Akizuki, K.~Yamaguchi, Y.~Morita, and Y.~Ohashi, ``The effect of proficiency level on measurement error of range of motion,'' \emph{Journal of physical therapy science}, vol.~28, no.~9, pp. 2644--2651, 2016.

\bibitem{bhatia2018consensus}
K.~P. Bhatia, P.~Bain, N.~Bajaj, R.~J. Elble, M.~Hallett, E.~D. Louis, J.~Raethjen, M.~Stamelou, C.~M. Testa, G.~Deuschl \emph{et~al.}, ``Consensus statement on the classification of tremors. from the task force on tremor of the international parkinson and movement disorder society,'' \emph{Movement disorders}, vol.~33, no.~1, pp. 75--87, 2018.

\end{thebibliography}
\end{document}